\documentclass[conference]{IEEEtran}
\IEEEoverridecommandlockouts

\usepackage{cite}
\usepackage{amsmath,amssymb,amsfonts}
\usepackage{algorithmic}
\usepackage{graphicx}
\usepackage{textcomp}
\usepackage{xcolor}

\usepackage{amssymb}
\usepackage{amsfonts}

\usepackage{booktabs}
\usepackage{multirow}

\def\BibTeX{{\rm B\kern-.05em{\sc i\kern-.025em b}\kern-.08em
    T\kern-.1667em\lower.7ex\hbox{E}\kern-.125emX}}
\begin{document}

\title{Sparse Hybrid Linear-Morphological Networks}


\author{Konstantinos Fotopoulos$^{1,2}$, Christos Garoufis$^1$, Petros Maragos$^{1,2}$\\
\small $^1$School of ECE, National Technical University of Athens, Athens, Greece\\\small $^2$Robotics Institute, Athena Research Center, Athens, Greece \\
{\tt\footnotesize k.fotopoulos@athenarc.gr, cgaroufis@mail.ntua.gr, maragos@cs.ntua.gr}
}

\maketitle

\begin{abstract}
We investigate hybrid linear-morphological networks. Recent studies highlight the inherent affinity of morphological layers to pruning, but also their difficulty in training. We propose a hybrid network structure, wherein morphological layers are inserted between the linear layers of the network, in place of activation functions. We experiment with the following morphological layers: 1) maxout pooling layers (as a special case of a morphological layer), 2) fully connected dense morphological layers, and 3) a novel, sparsely initialized variant of (2). We conduct experiments on the Magna-Tag-A-Tune (music auto-tagging) and CIFAR-10 (image classification) datasets, replacing the linear classification heads of state-of-the-art convolutional network architectures with our proposed network structure for the various morphological layers. We demonstrate that these networks induce sparsity to their linear layers, making them more prunable under L1 unstructured pruning. We also show that on MTAT our proposed sparsely initialized layer achieves slightly better performance than ReLU, maxout, and densely initialized max-plus layers, and exhibits faster initial convergence. 
\end{abstract}


\section{Introduction}

Deep neural networks (DNNs) achieve state-of-the-art performance in domains like computer vision, natural language processing, and audio understanding. Their success stems from their ability to learn hierarchical representations from large datasets. However, as DNNs grow in size, they demand significant computational and memory resources, making deployment on edge devices challenging \cite{8763885}. This has driven research into efficient architectures, particularly focusing on sparsity and pruning techniques \cite{MLSYS2020_6c44dc73}.

Pruning techniques aim to remove redundant weights from a network while preserving its predictive accuracy. Various methods, such as structured and unstructured pruning, weight quantization, and low-rank approximations, have been explored to enhance efficiency. L1 unstructured pruning \cite{li2016pruning}, for instance, removes weights based on their magnitude, leading to sparser models that require fewer computations. Despite the promise of such methods, designing inherently sparse neural networks remains an open challenge, as standard architectures rely heavily on dense weight matrices.


Mathematical morphology \cite{Heij94,Mara05b,NaTa10,Serr82} is a powerful set- and lattice-theoretic methodology, and provides a large variety of efficient nonlinear signal operators which have been widely used in image/signal processing and computer vision for many tasks including non-Gaussian denoising, connected filters, geometric feature extraction, representation, shape analysis, segmentation, remote sensing, object detection and recognition \cite{HaSh92,Heij94,Mara05b,Meye19,NaTa10,SaWi09,Serr82,Soil04}.
Its arithmetic is based on max-plus and min-plus operations, called dilations and erosions, and is
 closely related to tropical algebra \cite{Cuni79,Butk10}. These operations form the basis for morphological neural networks \cite{RitSus96,RiUr03,PeMa00}, which leverage dilations and erosions instead of standard linear inner products (i.e. multiply-accumulate operations). 
However, integrating these max/min operations into neural net learning pipelines has been challenging due to their non-differentiability \cite{PeMa00,FFY20,GDG23}.

A key property of morphological layers is their high affinity to pruning. Since they emphasize extremal values rather than summations over many parameters, they inherently favor sparsity. Studies \cite{DiMa21, Zha+19,GDG23} show that morphological networks are able to maintain accuracy at significantly lower parameter counts. However, their non-differentiability and lack of tailored optimization techniques make training difficult, limiting their practical use. To address this, we propose a hybrid linear-morphological structure, wherein the activation functions, typically intertwined with fully connected linear layers, are replaced with explicitly sparse morphological layers. This design balances trainability and pruning.


\textbf{Contributions:}\ 
Our contributions can be summarized as follows: 
1) We propose replacing the ReLU activation layers of linear networks and/or the pooling layers of maxout networks by explicitly sparse morphological layers, creating a hybrid linear-morphological neural network that balances trainability and prunability. 2) By replacing the linear classification head of state-of-the-art convolutional networks~\cite{won2020evaluation} with the aforementioned network structure, and evaluating our approach on the Magna-Tag-A-Tune~\cite{law2009evaluation} (MTAT) and CIFAR-10~\cite{krizhevsky2009learning} datasets, i) we demonstrate that our proposed network is competitive with ReLU-based networks and improves upon maxout and max-plus block networks~\cite{Zha+19} (in fact, it achieves the best performance on MTAT) 
while also exhibiting faster initial convergence, and ii) we show that 
our explicitly sparse max-plus block networks
naturally induce sparsity in their linear layers, making them significantly more prunable under L1 unstructured pruning.

\textbf{Related work:}\  
Various works have explored morphological networks and their applications with varying levels of success. Some works have focused on replacing the max-pooling operations of convolutional neural networks with morphological operations \cite{FFY20}. Others have introduced hybrid linear-morphological architectures, with the aim of alleviating the problem of their training \cite{MSMC22, Zha+19, Vall20}.

The most closely related works are \cite{Zha+19, DiMa21}. \cite{Zha+19} introduces the max-plus block—a combination of a linear and morphological layer. They show performance gains over \cite{ChMa17} and suggest potential for pruning. Our approach also forms max-plus blocks but differs in key ways: 1) Our morphological layers are explicitly constrained to being sparse, which aids in training and pruning. 2) We use a different topology; instead of replacing linear layers to keep parameter count constant—hindering test performance—we insert additional morphological layers in place of activations and/or maxout pooling, maintaining model performance. Unlike \cite{Zha+19}, which uses transfer learning on CIFAR, we train successfully from scratch.  3) \cite{Zha+19} prunes the weights only of the morphological layer, thereby indirectly deactivating only some linear neurons, effectively proving that the morphological layers are prunable. In contrast, we prune primarily the weights of the linear layers, showing that the linear layers themselves become more prunable, i.e. the inclusion of morphological layers induces sparsity to the linear layers themselves. 

\cite{DiMa21} focuses primarily on pruning, demonstrating that morphological layers are highly prunable, disregarding performance. We, on the other hand, i) have given emphasis to the performance of the networks, by assuming a different topology and sparsely initializing the morphological layers, and ii) demonstrate that the inclusion of morphological layers makes the rest of the layers more prunable.

\section{Preliminaries}

Tropical (minmax) algebra studies the tropical semirings, encompassing both the max-plus and min-plus semirings \cite{Butk10,Cuni79,MaSt15,MCT21}. The max-plus semiring \((\mathbb{R}_{\mathrm{max}}, \vee, +)\) is the set \(\mathbb{R}_{\mathrm{max}}=\mathbb{R}\cup \{-\infty\}\) equipped with the binary operations \(\vee\) (maximum), and \(+\) (ordinary addition), while the min-plus semiring \((\mathbb{R}_{\mathrm{min}}, \wedge, + )\) is the set \(\mathbb{R}_{\mathrm{min}}=\mathbb{R}\cup \{+\infty\}\) equipped with the binary operations \(\wedge\) (minimum) and \(+\). 
Within tropical algebra we can define matrix operations. For example, for compatible matrices $\mathbf{A},\mathbf{B}$, their max-plus  multiplication \(\boxplus\) is defined by \((\mathbf{A} \boxplus \mathbf{B})_{ij}=\bigvee_{k} a_{ik} + b_{kj}\), and their min-plus multiplication \(\boxplus'\) is defined by \((\mathbf{A} \boxplus' \mathbf{B})_{ij} = \bigwedge_{k} a_{ik} + b_{kj}\). 

Mathematical morphology is well-defined on complete lattices \cite{Heij94}, i.e. partially ordered sets in which every subset has a supremum and an infimum. Morphological operations map vectors and signals between complete lattices using two fundamental transformations: dilations (that distribute over suprema) and erosions (that distribute over infima). Shift-invariant signal dilations (erosions) correspond to nonlinear max-plus (min-plus) convolutions. 
These max-plus and min-plus operations connect mathematical morphology with tropical algebra.

In this paper, we focus on dilations and erosions defined on the set  \(\overline{\mathbb{R}}^n\) of finite discrete-time signals (i.e. vectors $\mathbf{x}=[x_i]$), where $\overline{\mathbb{R}}=\mathbb{R}\cup\{\pm \infty\}$. This  forms a complete lattice when equipped with the partial order \(\mathbf{x}\preceq \mathbf{y} \Leftrightarrow x_i\leq y_i, \forall i\in [n]\). 
For given weight vectors  \(\mathbf{w}, \mathbf{m} \in \mathbb{R}^n\) a dilation \(\delta_{\mathbf{w}}\) and erosion \(\varepsilon_{\mathbf{m}}\) from \(\overline{\mathbb{R}}^n\) to \(\overline{\mathbb{R}}\) can be defined as follows: 
\(
\delta_{\mathbf{w}}(\mathbf{x}) = \bigvee_{i\in[n]} (x_i+w_i) = \mathbf{w}^\top \boxplus \mathbf{x}
\)
and \(
\varepsilon_{\mathbf{m}}(\mathbf{x}) = \bigwedge_{i\in [n]} (x_i + m_i) = \mathbf{m}^\top \boxplus' \mathbf{x}.
\)
The weight vectors can be viewed geometrically as `structuring elements'. 
A max-plus Morphological Perceptron (MP) \cite{RiUr03} is simply a biased vector dilation, i.e. \(\mathrm{MP}_{\mathbf{w}}(\mathbf{x})=w_0\vee \delta_{\mathbf{w}[1:]}(\mathbf{x})=w_0 \vee \bigvee_{i\in [n]} (x_i + w_i)\).  
Similarly, a min-plus MP is  a biased vector erosion. These perceptrons resemble linear perceptrons, but with summation replaced by max or min and multiplication replaced by ordinary addition. Another variant, the Dilation-Erosion Perceptron (DEP), takes a convex combination of a dilation and an erosion and is trained using the Convex-Concave Procedure \cite{ChMa17}.

MPs can be treated as building blocks for the construction of more complex networks, termed as Morphological Neural Networks (MNNs) \cite{RitSus96,PeMa00}. A max-plus MP-based network is recursively defined as
\(
\mathbf{x}^{(n)}=f^{(n)} (\mathbf{w}^{(n)}_0\vee \mathbf{W}^{(n)} \boxplus \mathbf{x}^{(n-1)}), 
\)
where \(\mathbf{w}^{(n)}_0\) is the bias vector, \(\mathbf{W}^{(n)}\) is the weight matrix, and \(f^{(n)}\) is an activation function. 


\section{Proposed method}

\begin{figure}
    \centering
    \includegraphics[width=0.9\linewidth]{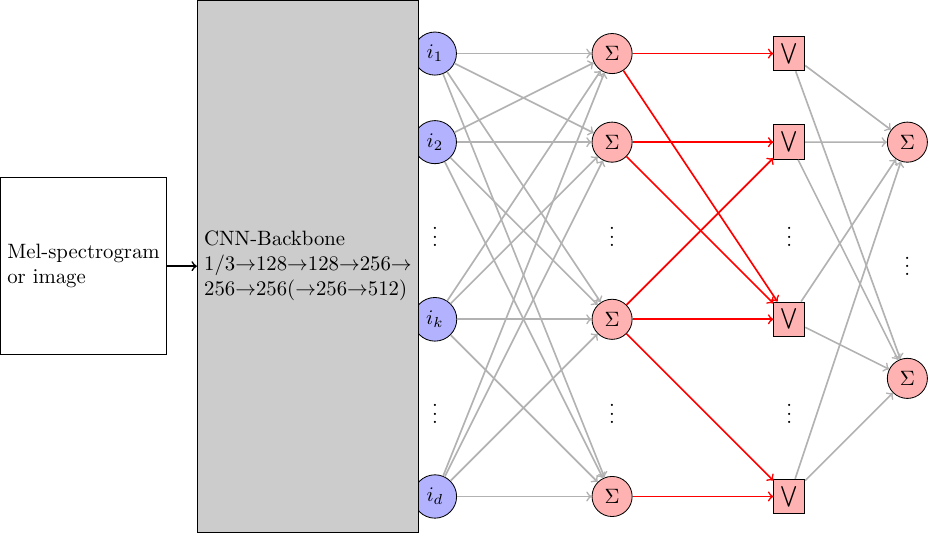}
    \caption{Proposed network structure, prepended by a CNN backbone, for the case of MTAT and CIFAR; ReLU activations have been replaced by a sparse morphological layer. Each MP has on average 2 input weights (red)}
    \label{fig:proposed_network}
    \vspace{-0.5cm}
\end{figure}

Before we proceed with our proposed method, we study ReLU activated and maxout networks, showing how these can be expressed 
by combining a linear and a morphological layer.  

\textbf{ReLU networks:}\ A fully connected ReLU activated network can be recursively defined as follows: 
\begin{equation}
\label{eq:relu}
\mathbf{x}^{(n)}=\max(\mathbf{A}^{(n)}\mathbf{x}^{(n-1)}+\mathbf{b}^{(n)}, \mathbf{0}),
\end{equation}
where the maximum operation is applied element-wise. 
We have: 
\[
x^{(n)}_i = 0 \vee ((\mathbf{A}^{(n)}\mathbf{x}^{(n-1)})_i + b^{(n)}_i)\vee \bigvee_{j\neq i}((\mathbf{A}^{(n)}\mathbf{x}^{(n-1)})_j -\infty). 
\]
We may write this network in the following equivalent form: 
\[
\mathbf{y}^{(n)} = \mathbf{A}^{(n)}\mathbf{x}^{(n-1)},
\]
\[
\mathbf{x}^{(n)} = \mathrm{diag}_{\mathrm{mp}}(b^{(n)}_i) \boxplus y^{(n)} \vee \mathbf{0} = \mathbf{W}^{(n)}_{\mathrm{ReLU}} \boxplus\mathbf{y}^{(n)}\vee \mathbf{0}, 
\]
where \(\mathrm{diag}_{\mathrm{mp}}\) denotes the max-plus diagonal matrix; i.e. off-diagonal elements equal to \(-\infty\). 
The above form shows that any biased ReLU activated fully connected linear layer can be written as an unbiased linear layer, followed by a zero-biased max-plus morphological layer with diagonal weight matrix. 

\textbf{Maxout networks:}\ A fully connected maxout network \cite{GWMB13} with a pooling of \(P\) and \(N_n\) maxout units in the \(n\)-th layer can be recursively defined as follows:
\[
\mathbf{x}^{(n)} = \max_{p\in [P]} (\mathbf{A}^{(n)}_p \mathbf{x}^{(n-1)} + \mathbf{b}^{(n)}_p), 
\]
where the maximum operation is applied element-wise. Define
\[
\mathbf{A}^{(n)}=\begin{bmatrix}
    (\mathbf{A}^{(n)}_1)^\top, 
    \cdots ,
    (\mathbf{A}^{(n)}_p)^\top
\end{bmatrix}^\top,
\]
\[
\mathbf{w}^{(n)} = \begin{bmatrix}
    (\mathbf{b}^{(n)}_1)^\top, 
    \cdots, 
    (\mathbf{b}^{(n)}_p)^\top
\end{bmatrix}^\top.
\]
Then, the network can equivalently be defined as follows: 
\[
\mathbf{g}^{(n)} = \mathbf{A}^{(n)}\mathbf{x}^{(n-1)} + \mathbf{w}^{(n)},
\]
\begin{equation}
\label{eq:maxout}
x^{(n)}_i = \max_{p\in [P]}(g^{(n)}_{i + (p-1)N_n}).
\end{equation}
This form shows that maxout networks are effectively a linear layer followed by max-pooling. Notice that we may write
\[
x^{(n)}_i = \max_{p\in [P]}((\mathbf{A}^{(n)}\mathbf{x}^{(n-1)})_{i+(p-1)N_n} + w^{(n)}_{i+(p-1)N_n})
\]
\[
= \max_{j = i+(p-1)N_n}((\mathbf{A}^{(n)}\mathbf{x}^{(n-1)})_{j} + w^{(n)}_{j}) 
\]
\[
\vee \max_{j \neq i+(p-1)N_n}((\mathbf{A}^{(n)}\mathbf{x}^{(n-1)})_{j} -\infty)
\]
Hence, we may write this network in the following form: 
\[
\mathbf{y}^{(n)} = \mathbf{A}^{(n)}\mathbf{x}^{(n-1)},
\]
\[
\mathbf{x}^{(n)}=\begin{pmatrix}
\mathbf{W}^{(n)}_1 & \mathbf{W}^{(n)}_2 & \cdots & \mathbf{W}^{(n)}_p
\end{pmatrix} \boxplus \mathbf{y}^{(n)} = \mathbf{W}^{(n)}_{\mathrm{maxout}} \boxplus \mathbf{y}^{(n)},
\]
where 
\[ \mathbf{W}^{(n)}_{k}=
\mathrm{diag}_{\mathrm{mp}}(b^{(n)}_{ki}). 
\]
The above form shows that any fully connected maxout linear layer can be written as an unbiased linear layer, followed by an unbiased max-plus morphological layer with a weight matrix formed by the concatenation of multiple diagonal matrices. 

\textbf{Max-plus block:}\ 
Introduced in \cite{Zha+19}, the max-plus block (with the inclusion of a morphological bias) generalizes the aforementioned structures by relaxing the constrained forms of the weight matrix and bias. Analytically, a network of max-plus blocks can be recursively defined as follows: 
\[
\mathbf{y}^{(n)} = \mathbf{A}^{(n)}\mathbf{x}^{(n-1)},
\]
\[
\mathbf{x}^{(n)} = \mathbf{W}^{(n)} \boxplus \mathbf{y}^{(n)} \vee \mathbf{w}^{(n)}_{0}.
\]
The weight matrices \(\mathbf{A}^{(n)}, \mathbf{W}^{(n)}\) and bias \(\mathbf{w}^{(n)}_0\) are general weight matrices and bias. From our previous analysis, we see that the max-plus block network is a generalization of ReLU and maxout networks, and hence it immediately follows that it is a universal approximator~\cite{Zha+19}. Notice that a max-plus block is effectively a maxout layer, where the linearities are shared across all outputs, achieving greater effective pooling. 

\textbf{Our method:}\ 
Our proposal is twofold: 1) We propose a new means of constructing hybrid linear-morphological topologies, which maintains test performance and is amendable to pruning, and, more importantly, 2) we propose a new constraint on the form of the max-plus block, where the morphological layer is explicitly defined to be sparse. 

Proposal 1): Most existing works focus on replacing linear layers with morphological layers, in order to keep parameter count constant. For example, in \cite{Zha+19}, the authors replace the final linear layer with a max-plus layer, turning a linear-ReLU-linear topology to  a linear-ReLU-morphological one. By contrast, we leave the linear layers as are, and replace activations with additional morphological layers, turning linear-ReLU-linear topologies into linear-morphological-linear ones. Specifically, if we are given a linear-ReLU layer as defined in \eqref{eq:relu} with \(\mathrm{n_{out}}\) outputs, i.e. \(\mathbf{A}^{(n)}\in \mathbb{R}^{\mathrm{n_{out}}\times \mathrm{n_{in}}}\), then we get rid of the ReLU layer and the linear bias, and add a morphological layer with \(\mathbf{W}^{(n)}\in \mathbb{R}^{\mathrm{n_{out}}\times \mathrm{n_{out}}}, \mathbf{w}^{(n)}_0\in \mathbb{R}^{\mathrm{n_{out}}}\), forming a max-plus block. If we are given a maxout layer as defined in \eqref{eq:maxout} with \(\mathrm{n_{out}}\) outputs and pooling of \(P\), i.e. \(\mathbf{A}^{(n)}\in \mathbb{R}^{P\mathrm{n_{out}} \times \mathrm{n_{in}}}\), then we get rid of the maxout pooling layer and the linear bias, set a new linear layer with \(\widetilde{\mathbf{A}}^{(n)}\in \mathbb{R}^{\mathrm{n_{out}} \times \mathrm{n_{in}}}\) and no linear bias, and a new morphological layer with \(\mathbf{W}^{(n)}\in \mathbb{R}^{\mathrm{n_{out}}\times \mathrm{n_{out}}}, \mathbf{w}^{(n)}_0\in \mathbb{R}^{\mathrm{n_{out}}}\), forming a max-plus block. 

Proposal 2): The max-plus block of \cite{Zha+19} defines a dense morphological layer, where each output has an effective pooling equal to the dimension of the input of the morphological layer. 
Instead, we propose that the morphological layer of the max-plus block be explicitly sparse. Specifically, for a morphological layer with \(\mathrm{n_{out}}\) outputs and inputs (as defined in Proposal 1), we explicitly initialize all but \(P \cdot \mathrm{n_{out}}\) weights to \(-\infty\), which renders them inactive throughout training. \(P\) is the effective pooling we wish each output to have on average. This means that we have the same effective pooling as a maxout network, but with a reduced linear layer of size \(\mathrm{n_{out}}\) instead of \(P \cdot \mathrm{n_{out}}\) (see Proposal 1). In practice, we take \(P=2\). We should note that Proposal 2 eliminates the vast majority of additional parameters that Proposal 1 introduces. 

Our method is illustrated in Fig.~\ref{fig:proposed_network}, where the ReLU activations of a linear-ReLU-linear topology have been 
replaced by a sparse morphological layer with \(P=2\). On average, each MP has 2 active inputs; in contrast to maxout layers, the number of active inputs differs throughout each unit, whereas the same input can be fed into multiple units.

\section{Experiments}

\begin{table}[t]
    \begin{center}
        \vspace{-0.6cm}
        \caption{Test performance of 
        different methods on MTAT}
        \label{table:1}
        \vspace{-0.15cm}
        \begin{tabular}{lcc}
            \toprule
            Method & ROC-AUC $\uparrow$ & PR-AUC $\uparrow$ \\ 
            \midrule
            ReLU & 0.9149 $\pm$ 0.0005 & 0.4632 $\pm$ 0.0023 \\
            Maxout & 0.9148 $\pm$ 0.0003 & 0.4626 $\pm$ 0.0021 \\
            Zhang et al. \cite{Zha+19} & 0.5000 $\pm$ 0.0000 & 0.0653 $\pm$ 0.0000 \\
            Dense-Morph & 0.9127 $\pm$ 0.0013 & 0.4553 $\pm$ 0.0010 \\
            Sparse-Morph (ours) & \textbf{0.9152} $\pm$ 0.0002 & \textbf{0.4646} $\pm$ 0.0020 \\
            \bottomrule
        \end{tabular}
    \end{center}
\vspace{-0.8cm}
\end{table}

\begin{table}[t]
    \begin{center}
        \caption{Results on CIFAR-10}
        \label{table:cifar_pruning}
        \vspace{-0.35cm}
        \setlength{\tabcolsep}{0.3em}
        \resizebox{\columnwidth}{!}{
        \begin{tabular}{llc|cccc}
        \toprule
        \multicolumn{2}{c}{Pruning ratio} & \multicolumn{4}{c}{Test Accuracy (\%)} \\
        \cmidrule(lr){1-2} \cmidrule(lr){3-6}
        $r_2$ & $r_1$ (params) & ReLU & Maxout & Dense-Morph & Sparse-Morph \\
        \midrule
\multicolumn{2}{c}{Original} & \textbf{78.98} $\pm$ 0.76 & 78.17 $\pm$ 0.86 & 77.78 $\pm$ 0.72 & 78.32 $\pm$ 0.50 \\
0.7 & 0.7 (41380) & 69.57 $\pm$ 3.38 & 34.89 $\pm$ 6.62 & 29.83 $\pm$ 11.16 & \textbf{74.13} $\pm$ 2.14 \\
0.7 & 0.8 (28273) & 55.03 $\pm$ 6.49 & 23.25 $\pm$ 4.86 & 21.61 $\pm$ 8.35 & \textbf{66.10} $\pm$ 4.02 \\
0.7 & 0.9 (15166) & 27.78 $\pm$ 5.84 & 15.41 $\pm$ 5.24 & 17.14 $\pm$ 4.02 & \textbf{34.16} $\pm$ 10.41 \\
0.8 & 0.7 (40868) & 62.36 $\pm$ 3.83 & 34.42 $\pm$ 4.77 & 24.83 $\pm$ 9.63 & \textbf{72.40} $\pm$ 2.33 \\
0.8 & 0.8 (27761) & 43.33 $\pm$ 5.99 & 22.08 $\pm$ 3.79 & 19.66 $\pm$ 8.48 & \textbf{62.51} $\pm$ 4.60 \\
0.8 & 0.9 (14654) & 20.54 $\pm$ 6.58 & 14.24 $\pm$ 4.31 & 17.03 $\pm$ 6.46 & \textbf{30.73} $\pm$ 8.90 \\
0.9 & 0.7 (40356) & 44.77 $\pm$ 4.86 & 35.34 $\pm$ 3.23 & 19.61 $\pm$ 6.45 & \textbf{70.31} $\pm$ 3.02 \\
0.9 & 0.8 (27249) & 26.80 $\pm$ 4.56 & 23.76 $\pm$ 1.42 & 14.68 $\pm$ 4.34 & \textbf{58.20} $\pm$ 7.24 \\
0.9 & 0.9 (14142) & 14.91 $\pm$ 3.93 & 15.85 $\pm$ 3.78 & 13.93 $\pm$ 3.81 & \textbf{28.57} $\pm$ 12.26 \\
0.95 & 0.7 (40100) & 36.49 $\pm$ 1.45 & 32.21 $\pm$ 2.39 & 22.11 $\pm$ 7.63 & \textbf{67.36} $\pm$ 2.85 \\
0.95 & 0.8 (26993) & 22.53 $\pm$ 4.74 & 23.85 $\pm$ 2.78 & 17.42 $\pm$ 5.79 & \textbf{54.92} $\pm$ 5.93 \\
0.95 & 0.9 (13886) & 13.27 $\pm$ 3.04 & 16.87 $\pm$ 5.54 & 14.53 $\pm$ 3.92 & \textbf{25.88} $\pm$ 7.63 \\

        \bottomrule
        \end{tabular}}
    \vspace{-0.6cm}
    \end{center}
\end{table}

\begin{table*}[t]
    \begin{center}
        \caption{Pruning performance of different methods for a variety of pruning ratios \(r_1, r_2\) on MTAT}
        \label{table:2}
        \vspace{-0.4cm}
        \setlength{\tabcolsep}{0.5em}
        \resizebox{\textwidth}{!}{
        \begin{tabular}{llcc|cccccc}
         \toprule
        \multicolumn{2}{c}{Pruning ratio}& \multicolumn{2}{c}{ReLU} & \multicolumn{2}{c}{Maxout} & \multicolumn{2}{c}{Dense-Morph.} & \multicolumn{2}{c}{Sparse-Morph. (ours)} \\
        \cmidrule(lr){1-2} \cmidrule(lr){3-4} \cmidrule(lr){5-6} \cmidrule(lr){7-8} \cmidrule(lr){9-10}
        $r_2$ & $r_1$ (\#params) & ROC-AUC $\uparrow$ & PR-AUC $\uparrow$ & ROC-AUC $\uparrow$ & PR-AUC $\uparrow$ & ROC-AUC $\uparrow$ & PR-AUC $\uparrow$ & ROC-AUC $\uparrow$ & PR-AUC $\uparrow$ \\
        \midrule
        \multirow{4}{*}{0.8}  & 0.8 (58111)  & 0.9048 $\pm$ 0.0006 & 0.4278 $\pm$ 0.0014  & 0.9116 $\pm$ 0.0014 & 0.4530 $\pm$ 0.0022  & 0.9023 $\pm$ 0.0043 & 0.4304 $\pm$ 0.0046  & \textbf{0.9125} $\pm$ 0.0008 & \textbf{0.4585} $\pm$ 0.0026 \\
          & 0.9 (31897) & 0.9045 $\pm$ 0.0006 & 0.4247 $\pm$ 0.0026  & 0.9100 $\pm$ 0.0015 & 0.4501 $\pm$ 0.0023  & 0.9005 $\pm$ 0.0034 & 0.4232 $\pm$ 0.0043  & \textbf{0.9119} $\pm$ 0.0009 & \textbf{0.4563} $\pm$ 0.0022 \\
          & 0.95 (18790) & 0.9034 $\pm$ 0.0006 & 0.4212 $\pm$ 0.0047  & 0.9068 $\pm$ 0.0010 & 0.4440 $\pm$ 0.0021  & 0.8979 $\pm$ 0.0031 & 0.4171 $\pm$ 0.0045  & \textbf{0.9102} $\pm$ 0.0012 & \textbf{0.4514} $\pm$ 0.0035\\
          & 0.98 (10925) & 0.8995 $\pm$ 0.0011 & 0.4099 $\pm$ 0.0069  & 0.8988 $\pm$ 0.0015 & 0.4282 $\pm$ 0.0050  & 0.8939 $\pm$ 0.0034 & 0.4090 $\pm$ 0.0059  & \textbf{0.9058} $\pm$ 0.0019 & 
          \textbf{0.4397} $\pm$ 0.0052 \\
        \midrule
        \multirow{4}{*}{0.9}  & 0.8 (55551) & 0.8911 $\pm$ 0.0024 & 0.4039 $\pm$ 0.0049  & 0.9000 $\pm$ 0.0011 & 0.4466 $\pm$ 0.0024  & 0.8914 $\pm$ 0.0068 & 0.4128 $\pm$ 0.0072  & \textbf{0.9074} $\pm$ 0.0028 & \textbf{0.4480} $\pm$ 0.0039 \\
          & 0.9 (29337) & 0.8901 $\pm$ 0.0033 & 0.4001 $\pm$ 0.0072  & 0.8983 $\pm$ 0.0013 & 0.4427 $\pm$ 0.0028  & 0.8910 $\pm$ 0.0054 & 0.4096 $\pm$ 0.0053  & \textbf{0.9064} $\pm$ 0.0030 & \textbf{0.4454} $\pm$ 0.0033 \\
          & 0.95 (16230) & 0.8892 $\pm$ 0.0029 & 0.3961 $\pm$ 0.0082  & 0.8950 $\pm$ 0.0010 & 0.4363 $\pm$ 0.0020  & 0.8899 $\pm$ 0.0045 & 0.4067 $\pm$ 0.0045  & \textbf{0.9046} $\pm$ 0.0032 & \textbf{0.4401} $\pm$ 0.0047 \\
          & 0.98 (8365) & 0.8836 $\pm$ 0.0048 & 0.3807 $\pm$ 0.0106  & 0.8885 $\pm$ 0.0011 & 0.4214 $\pm$ 0.0036  & 0.8871 $\pm$ 0.0045 & 0.4004 $\pm$ 0.0061  & \textbf{0.8993} $\pm$ 0.0041 & \textbf{0.4258} $\pm$ 0.0040 \\
        \midrule
        \multirow{4}{*}{0.95}  & 0.8 (54271) & 0.8647 $\pm$ 0.0017 & 0.3630 $\pm$ 0.0034  & 0.8943 $\pm$ 0.0011 & \textbf{0.4392} $\pm$ 0.0030  & 0.8730 $\pm$ 0.0083 & 0.3938 $\pm$ 0.0087  & \textbf{0.8965} $\pm$ 0.0058 & 0.4325 $\pm$ 0.0037 \\
          & 0.9 (28057) & 0.8627 $\pm$ 0.0041 & 0.3585 $\pm$ 0.0060  & 0.8923 $\pm$ 0.0009 & \textbf{0.4344} $\pm$ 0.0035  & 0.8730 $\pm$ 0.0064 & 0.3919 $\pm$ 0.0071  & \textbf{0.8952} $\pm$ 0.0061 & 0.4292 $\pm$ 0.0033 \\
          & 0.95 (14950) & 0.8602 $\pm$ 0.0047 & 0.3477 $\pm$ 0.0092  & 0.8882 $\pm$ 0.0013 & \textbf{0.4271} $\pm$ 0.0047  & 0.8725 $\pm$ 0.0052 & 0.3892 $\pm$ 0.0059  & \textbf{0.8930} $\pm$ 0.0066 & 0.4223 $\pm$ 0.0039 \\
          & 0.98 (7085) & 0.8506 $\pm$ 0.0061 & 0.3289 $\pm$ 0.0150  & 0.8789 $\pm$ 0.0019 & \textbf{0.4072} $\pm$ 0.0042  & 0.8692 $\pm$ 0.0037 & 0.3831 $\pm$ 0.0051  & \textbf{0.8866} $\pm$ 0.0074 & 0.4040 $\pm$ 0.0039 \\
        \midrule
        \multirow{4}{*}{0.98}  & 0.8 (53503) & 0.7820 $\pm$ 0.0083 & 0.2732 $\pm$ 0.0044  & 0.8398 $\pm$ 0.0101 & 0.3919 $\pm$ 0.0086  & 0.8058 $\pm$ 0.0129 & 0.3401 $\pm$ 0.0149  & \textbf{0.8737} $\pm$ 0.0072 & \textbf{0.3951} $\pm$ 0.0086 \\
          & 0.9 (27289) & 0.7797 $\pm$ 0.0092 & 0.2680 $\pm$ 0.0063  & 0.8361 $\pm$ 0.0100 & 0.3836 $\pm$ 0.0057  & 0.8061 $\pm$ 0.0120 & 0.3402 $\pm$ 0.0114  & \textbf{0.8719} $\pm$ 0.0077 & \textbf{0.3917} $\pm$ 0.0088 \\
          & 0.95 (14182) & 0.7755 $\pm$ 0.0069 & 0.2606 $\pm$ 0.0119  & 0.8302 $\pm$ 0.0088 & 0.3736 $\pm$ 0.0063  & 0.8060 $\pm$ 0.0109 & 0.3382 $\pm$ 0.0088  & \textbf{0.8683} $\pm$ 0.0081 & \textbf{0.3807} $\pm$ 0.0080 \\
          & 0.98 (6317) & 0.7695 $\pm$ 0.0057 & 0.2479 $\pm$ 0.0093  & 0.8147 $\pm$ 0.0094 & 0.3462 $\pm$ 0.0095  & 0.8033 $\pm$ 0.0103 & 0.3317 $\pm$ 0.0063  & \textbf{0.8584} $\pm$ 0.0084 & \textbf{0.3565} $\pm$ 0.0084 \\
        \bottomrule
        \end{tabular}}
    \end{center}
    \vspace{-0.6cm}
\end{table*}


\textbf{Experimental Setup:}\ 
Our goal is threefold: 1) to ensure that including morphological layers, particularly sparse ones, in state-of-the-art architectures does not degrade test performance, 2) to show that they contribute to accelerated training, 
and 3) to examine whether maxout, max-plus block, and our sparse max-plus block networks improve network prunability.

We primarily conduct experiments on the MTAT dataset~\cite{law2009evaluation}, a widely used benchmark in music tagging~\cite{won2020evaluation}, containing 25,863 annotated 29-sec song excerpts. Following prior work~\cite{lee18}, we evaluate on the top 50 tags using the default splits. We also evaluate our method on CIFAR-10~\cite{krizhevsky2009learning}, which consists of 60,000 32 $\times$ 32 grayscale images, distributed evenly into 10 classes.

Our base model is the short-chunk CNN~\cite{won2020evaluation}, which processes log-mel spectrograms\footnote{3.69 sec length -- 96 mel bands -- 512-sample windows, 256-sample hop} and consists of a 7-layer convolutional backbone followed by a two-layer 
linear-ReLU-linear classification head; we replace the classification head with a linear-morphological-linear design. The first FC layer (512 neurons) receives a 512-dimensional input, and the final layer has 50 output neurons. The intermediate morphological layer has size $512\times512$ with mostly inactive parameters. For CIFAR experiments, we adapt the convolutional backbone by reducing the number of convolutional layers to five.

We compare five classification heads: 1) ReLU-based MLP – a standard linear-ReLU-linear structure as in \cite{won2020evaluation}, 2) Maxout-based MLP – replacing the first ReLU layer with a maxout layer (pooling factor \(P=2\)), 3) Zhang et al. \cite{Zha+19} - replacing the final linear layer with a morphological layer, as proposed in \cite{Zha+19} for their CIFAR architecture, 
4) Dense-Morph-based MLP, replacing the first linear-ReLU layer with a dense linear-morphological layer (Proposal 1), and 5) Sparse-Morph-based MLP, which further constrains the morphological layer to be sparse (Proposals 1 \& 2). We incorporated Batch Normalization in all classification heads, with the exception of (4) on MTAT, where its removal led to smoother training.

All models are trained from scratch. For CIFAR-10 we train using Adam for 10 epochs with a learning rate of 0.001 and a random 80-20 train-validation split. For MTAT we train for 100 epochs. We use Adam \cite{kingma2015adam} with a learning rate of 0.0001 for the first 80 epochs, followed by SGD with Nesterov momentum (0.9) and a learning rate of 0.001 for 20 epochs. Weight decay is set to 0.0001. The model with the lowest validation loss is selected for testing. We train 5 models for each method, reporting the mean and standard error of the ROC-AUC, PR-AUC scores (MTAT) or classification accuracy (CIFAR-10) as measured in the testing split; for MTAT, excerpt-wise scores are obtained by averaging the per-spectrogram network outputs. 






\textbf{Test Performance:}\ Since morphological layers are known to be difficult to train, we first verify that their inclusion does not significantly degrade test accuracy. Thus, we report on the results of fully-trained, non-pruned networks in MTAT. These results are presented in Table~\ref{table:1}, showing that models that follow Proposal 1 achieve comparable results to ReLU and Maxout networks. In addition, the model following both Proposals 1 and 2, i.e. our Sparse-Morph-based, marginally achieves the best performance out of all the models. 



\textbf{Pruning Experiments:}\ We evaluate prunability via unstructured L1 pruning on the FC layers. Networks following Proposal 1 are pruned based on L1 norm (linear layers) or absolute magnitude (morphological layers, with pruned weights set to \(-\infty\)).
We prune the last linear layer with a pruning ratio of \(r_2\) and the remaining layers of the classification head with \(r_1\). To ensure equal parameter counts, we make the following adjustments: 1) For the maxout network, whose linear layer is twice the size of the ReLU network, we prune at a pruning ratio \(r_1'=1-(1-r_1)/2\), and an additional \(512\) parameters due to the biases. 2) For the Dense-Morph network, which has 1 linear and 1 morphological FC layer, we prune each at a pruning ratio \(r_1'=1-(1-r_1)/2\). 3) For the Sparse-Morph network, which has \(2 \cdot 512\) additional parameters, we prune this many additional parameters from the first linear layer. 

\begin{figure}
    \centering
    \includegraphics[width=0.6\linewidth]{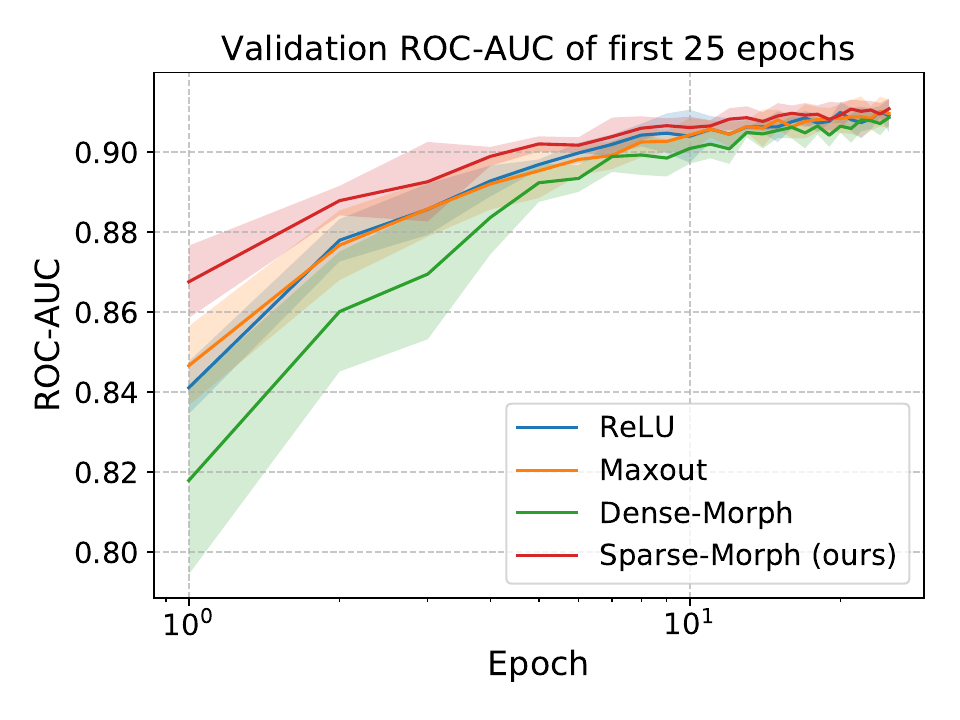}
    \includegraphics[width=0.6\linewidth]{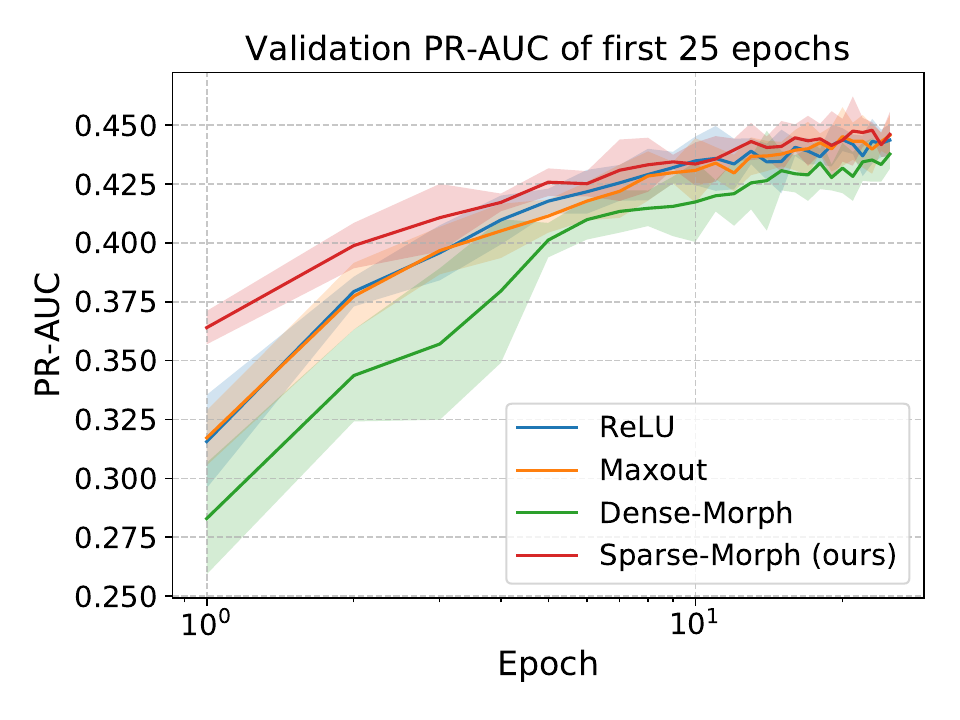}
    \vspace{-0.4cm}
    \caption{Error plot of validation ROC-AUC and PR-AUC scores of different methods on MTAT for the first 25 epochs of training; 0.9 confidence interval, x-axis is in log-scale.}
    \label{fig:convergence}
    \vspace{-0.5cm}
\end{figure}

Tables~\ref{table:cifar_pruning} \&~\ref{table:2} show that our sparse max-plus block networks achieve significantly better pruning performance than ReLU networks. This also holds true for the Maxout and Dense-Morph networks in MTAT (where Sparse-Morph network performs best in ROC-AUC, while Maxout and Sparse-Morph achieve similar PR-AUC scores), as well as in CIFAR under high pruning ratios; for other ratios, ReLU outperforms other morphological variants. Since
weights are mostly (in fact, for the Maxout and Sparse-Morph networks, solely) pruned from the preceding linear layers, it is implied that morphological layers are not only sparse themselves but also induce sparsity in adjacent linear layers.  


\textbf{Speed of Convergence:}\ Lastly, we compare the initial convergence speed of our models (Fig.~\ref{fig:convergence}) in MTAT. We observe that our Sparse-Morph model, which replaces activations with sparse morphological layers, achieves faster validation score improvements in the first 25 epochs; in contrast, Dense-Morph is slower to converge than the ReLU-based baseline.

\section{Conclusion}
We explored hybrid linear-morphological architectures and introduced a method to integrate max-plus blocks while preserving test performance. By replacing activation layers with morphological layers and sparsifying them via \(-\infty\) initialization, we achieved the best test performance in MTAT, fastest convergence, and enhanced prunability. Notably, we show that morphological layers induce sparsity in linear components, making hybrid networks—including maxout and max-plus variants—more prunable. Our findings highlight the potential of morphological layers for improving efficiency, convergence, and compression in neural networks.

\bibliographystyle{plain}
%
\bibliography{refs}

\end{document}